\def\year{2020}\relax
\documentclass[letterpaper]{article} %
\usepackage{aaai20}  %
\usepackage{times}  %
\usepackage{helvet} %
\usepackage{courier}  %
\usepackage[hyphens]{url}  %
\usepackage{graphicx} %
\usepackage{graphicx}  %
\nocopyright
\title{Monte Carlo Anti-Differentiation for\\Approximate Weighted Model Integration}
\author{Pedro Zuidberg Dos Martires \and Samuel Kolb \\ %
KU Leuven \\
pedro.zudo@kuleuven.be, samuel.kolb@kuleuven.be %
}

\usepackage{tikz}
\usetikzlibrary{shapes}

\usepackage{xcolor}

\usepackage{amsmath}
\usepackage{amsfonts}
\usepackage{amsthm}
\usepackage{nicefrac}
\usepackage{textgreek}
\usepackage{dsfont}
\usepackage{txfonts}

\usepackage{algorithm, algpseudocode}
\usepackage{mdframed}
\usepackage{subcaption}

\usepackage{microtype} 

\usepackage[shortlabels]{enumitem}

\newenvironment{talign}
 {\align}
 {\endalign}

\newtheorem{definition}{Definition}

\newtheorem{example}{Example}

\newtheorem{theo}{Theorem}

\newtheorem{proposition}{Proposition}

\usepackage{xspace}

\newcommand{\wmi}{\ensuremath{\mathrm{WMI}}\xspace}

\newcommand{\support}{\ensuremath{\phi}\xspace}
\newcommand{\wsupport}{\ensuremath{\phi}\xspace}
\newcommand{\weight}{\ensuremath{w}\xspace}
\newcommand{\wweight}{\ensuremath{\omega}\xspace}
\newcommand{\fso}{\ensuremath{\wvol}}

\newcommand{\ive}[1]{\llbracket#1\rrbracket}
\newcommand{\varset}[1]{\ensuremath{\mathbf{#1}}}

\newcommand{\xvars}{\varset{x}}
\newcommand{\bvars}{\varset{b}}

\newcommand{\wvol}{\ensuremath{\mathrm{vol}}\xspace}
\newcommand{\lra}{\ensuremath{\mathcal{LRA}}\xspace}
\newcommand{\lsmt}{\textlambda-SMT\xspace}

\newcommand{\polyvol}{\ensuremath{v}\xspace}

\newcommand{\fxsdd}{F-XSDD\xspace}
\newcommand{\fmcad}{F-XSDD(\textsc{Mcad})\xspace}

\newcommand{\latte}{\texttt{LattE Integrale}\xspace}
\newcommand{\vinci}{\texttt{VINCI}\xspace}
\newcommand{\volesti}{\texttt{VolEsti}\xspace}
\newcommand{\cpp}{\texttt{C++}\xspace}

\newcommand{\antid}{\ensuremath{\Psi}\xspace}

\newcommand{\bvec}[1]{{\bf #1}\xspace}

\begin{document}

\maketitle

\begin{abstract}
Probabilistic inference in the hybrid domain, i.e. inference over discrete-continuous domains,
requires tackling two well known \#P-hard problems 1)~weighted model counting (WMC) over discrete variables and 2)~integration over continuous variables.
For both of these problems inference techniques have been  developed separately in order to manage their \#P-hardness, such as knowledge compilation for WMC and Monte Carlo (MC) methods for (approximate) integration in the continuous domain.
Weighted model integration (WMI), the extension of WMC to the hybrid domain, has been proposed as a formalism to study probabilistic inference over discrete and continuous variables alike.
Recently developed WMI solvers have focused on exploiting structure in WMI problems, for which they rely on symbolic integration to find the primitive of an integrand, i.e. to perform anti-differentiation.
To combine these advances with state-of-the-art Monte Carlo integration techniques, we introduce \textit{Monte Carlo anti-differentiation} (MCAD), which computes MC approximations of anti-derivatives.
In our empirical evaluation we substitute the exact symbolic integration backend in an existing WMI solver with an MCAD backend. Our experiments show that that equipping existing WMI solvers with MCAD yields a fast yet reliable approximate inference scheme.
\end{abstract}

\section{Introduction}

At the heart of probabilistic AI lies the problem of performing probabilistic inference, which is a \#P-hard problem. For discrete random variables, the reduction to weighted model counting (WMC)~\cite{darwiche2009modeling} has emerged as the go-to technique to manage the hardness of probabilistic inference by exploiting structure, such as determinism and context-specific independence~\cite{chavira2008probabilistic}. Weighted Model Integration (WMI)~\cite{belle2015probabilistic} extends the WMC task from the discrete domain to the continuous domain by allowing for continuous random variables.

\begin{figure}[t!]
    \includegraphics[width=0.95\linewidth]{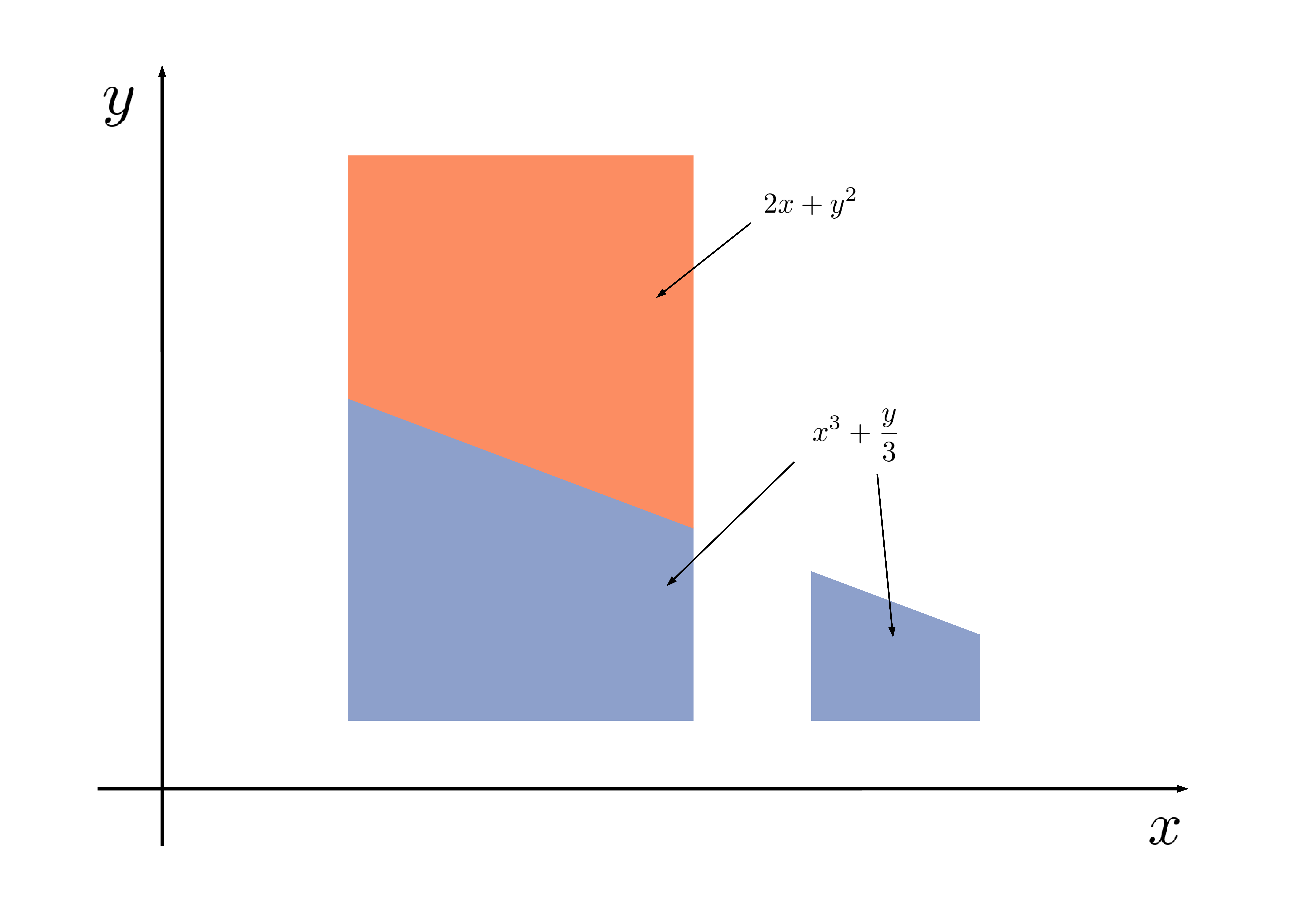} 
    \caption{Example of a WMI problem.}
    \label{figure:intro_example_wmi}
\end{figure}

\begin{example}
Consider the example of a WMI problem in Figure~\ref{figure:intro_example_wmi}. The problem has two continuous random variables ($x$ and $y$) and three Boolean random variables which produce the different feasible regions (the red region and the two blue regions). The regions themselves are given by constraints on the continuous variables. Moreover, for each feasible region a weight function is given. Outside of the regions the weight is zero. WMI tackles the problem of computing the integral over the feasible regions.
\end{example}

Recently, it has been shown~\cite{kolb2018efficient,kolb2019structure,zeng2019efficient} that, just as for WMC, exploiting structure in highly structured WMI problems is possible and can in practice lead to exponential-to-polynomial speed-ups in inference time. In contrast to WMC, however, WMI exhibits one major complication: integrating out continuous random variables when performing probabilistic inference. WMI solvers capable of exploiting structure in WMI problems,
such  as the \fxsdd family of algorithms~\cite{kolb2019structure},
rely on symbolic probabilistic inference, i.e. performing integrals symbolically. Symbolic integration is the problem of performing anti-differentiation (finding the indefinite integral also called the primitive).

Computing integrals (definite as well as indefinite) is a computationally hard problem, in the case of integrating polynomials, for example, \text{\#P-hard}~\cite{valiant1979complexity}, i.e. the number of functions calls necessary to compute an integral grows exponentially as the dimensionality increases. This dependency of the complexity class on the dimensionality of a problem is also referred to as \textit{curse of dimensionality}\footnote{The term \textit{curse of dimensionality} was originally coined in~\cite{bellman1957dynamic} in the context of dynamic programming but has since also been applied to the problem of integration.}. A popular technique, capable of circumventing the curse of dimensionality for definite integrals, is Monte Carlo (MC) integration.

This brings us to the crux of this paper. On the one hand, we would like to {\bf exploit structure present in WMI problems}, for which we need to be able to calculate indefinite integrals. On the other hand, we want to {\bf circumvent the curse of dimensionality of integration} by using MC techniques. 
Unfortunately, vanilla MC integration is not capable of calculating indefinite integrals. Therefore, we introduce the concept of Monte Carlo anti-differentiation (MCAD) in order to {\bf approximately compute the anti-derivative of an integrand with a Monte Carlo estimate}.

For the empirical evaluation we integrated MCAD into the existing \fxsdd family of WMI solvers where we use MCAD as a drop-in replacement for the exact symbolic integration backend. Experimentally, we show that approximating the anti-derivative with a Monte Carlo estimate, instead of computing symbolic indefinite integrals, allows for efficiently solving highly-structured WMI problems in high-dimensional domains --- yielding a practical WMI solver.

\section{Preliminaries}\label{sec:preliminaries}

\subsection{Weighted Model Integration}\label{sec:wmi}
WMI is the extension of WMC from propositional logic formulas to so-called {\em satisfiability modulo theory} (SMT) formulas. An SMT formula is a first-order formula with respect to a decidable background theory. Following~\cite{morettin2017efficient}, we define SMT($\lra$) formulas, i.e. SMT formulas that use {\em linear real arithmetics} as background theories
:
\begin{definition}\label{def:smtlra} 
(SMT($\lra$)) Let $\bvars$ be a set of $M$ Boolean and $\xvars$ a set of $N$ real variables. An atomic formula is an expression of the form $\sum_i c_i \cdot x_i {\bowtie} c$, where the $x_i \in \xvars$ and $c_i,c \in \mathbb{Q}$, and ${\bowtie} \xspace{\in} \{ =, \neq, \geq,\leq,>,< \}$.
We then define SMT($\lra$) theories as Boolean combinations (by means of the standard Boolean operators $\{\neg, \land, \lor, \rightarrow, \leftrightarrow \}$) of Boolean variables $b_i \in \bvars$ and of atomic formulas over $\xvars$.
\end{definition}

Following Equation~$5$ in~\cite{kolb2019structure}, we define weighted model integration in function of an indicator function over an SMT($\lra$) formula.
\begin{definition}(WMI)\label{def:wmi}
Given a set $\bvars$ of $M$ Boolean variables,  $\xvars$ of $N$ real variables, a weight function $\weight: \mathbb{B}^M \times \mathbb{R}^N \rightarrow \mathbb{R}_{\geq 0}$, and a support $\support$, in the form of an SMT formula, over $\bvars$ and  $\xvars$, the {\bf weighted model integral} is given by:
\begin{talign}
  \wmi(\support,\weight {\mid} \xvars, \bvars) = \sum_{\bvars} \int \ive{\support(\xvars, \bvars)} \weight({\xvars},{\bvars}) d{\xvars} \label{eqn:wmi_iverson_weight}
\end{talign}
where we use the Iverson bracket notation in $\ive{\support(\xvars, \bvars)}$ to denote the indicator function of $\support(\xvars, \bvars)$.
\end{definition}
In~\cite[Equation 8]{kolb2019structure}, the authors further manipulate the expression for computing the weighted model integral into a summation of weighted model integrals where the weight function does not dependent on $\bvars$ anymore:
\begin{talign}
  \sum_{\bvars} \int \ive{\support(\xvars, \bvars)} \weight({\xvars},{\bvars}) d{\xvars}&=
  \sum_{i}
  \underbrace{\sum_{\bvars} \int \ive{\wsupport^\bvars_i(\xvars)} \wweight_i(\xvars) d{\xvars}
  }_{\eqqcolon \wvol(\wsupport_i,\wweight_i|\xvars,\bvars)}\label{eqn:volume_defin}
\end{talign}

Solving a WMI problem over an SMT($\lra$) formula can hence be reduced to a two step procedure: 1)~rewriting the problem into a sum over tuples~$\langle \wsupport_i, \wweight_i \rangle$ of disjoint convex polytopes and weight functions, and 2)~integrating every weight~$\wweight_i$ over the corresponding support~$\wsupport_i$. The WMI is obtained by summing up the results obtained in the second step.
The first step of this procedure was coined {\bf \emph{\lsmt}}~\cite[Defintion 6]{kolb2019structure}. The \lsmt problem lies at the heart of all WMI solvers. Efficient WMI solvers are characterized by efficiently solving the \lsmt problem by exploiting redundancies in the SMT($\lra$) formula, thereby avoiding the computation of superfluous integrals. Solving the \lsmt problem and the integration step are the two \#P-hard problems that characterize  probabilistic inference in the hybrid domain.

\subsection{Exact Integration Techniques for WMI}\label{sec:symbolic_integration}
Techniques to compute integrals can be split into two categories 1) symbolic integration and 2) numerical integration. Symbolic integration is the problem of finding the anti-derivative, or indefinite integral, used to compute exact definite integrals, whereas numerical integration is a family of algorithms for calculating definite integrals. Numerical integration algorithms are either exact, e.g.~\cite{bueler2000exact,de2004effective}, or approximate, e.g.~\cite{metropolis1953equation,duane1987hybrid}. In the remainder of this subsection we give an overview of algorithms used so far in the WMI literature and discuss their advantages and shortcomings\footnote{We limit the discussion to approaches that handle general WMI problems over SMT(\lra) formulas, in contrast to~\cite{belle2016component} and \cite{zeng2019efficient}.}.

{\bf Fourier-Motzkin Elimination} What Gaussian elimination is to systems of strict linear equalities, is Fourier-Motzkin elimination~\cite{imbert1990redundant} to systems of linear inequalities. First and foremost Fourier-Motzkin elimination is a method to solve a system of linear inequalities but does also underlie a number of symbolic integration algorithms~\cite{sanner2011symbolic,gehr2016psi,kolb2018efficient} (even though this not being mentioned explicitly). As an underlying component of symbolic integration methods, the Fourier-Motkin elimination has also found its way into WMI solvers~\cite{kolb2018efficient,zuidbergdosmartires2018exactcompilation,kolb2019structure}.

{\bf Decomposition into Simplices} An other strategy for exact integration of polynomials over convex polytopes is decomposing a polytope into (signed) simplices and adding up the (signed) volumes of the simplices. \vinci~\cite{bueler2000exact} and  \latte~\cite{de2004effective,de2013software} are realization of such decomposition algorithms.
Even though \latte is technically a \textit{symbolic} algorithm, we consider it a numeric integration algorithm as it can only produce definite integrals. \latte has been used in WMI solvers of~\cite{belle2015probabilistic,belle2015hashing,morettin2017efficient,morettin2019advanced}.

\subsection{MCMC Volume Estimation of  Convex Bodies}\label{sec:volesti}
Monte Carlo integration is a popular technique to escape the curse of dimensionality for computing integrals or to approximate integrals for which no analytic solution is available and which hence cannot be computed symbolically. The simplest MC strategy is naive rejection sampling. Even though rejection sampling is straightforward to implement, it hardly circumvents the curse of dimensionality: for high-dimensional spaces the rejection rate is prohibitively high.

To repair this problem, a plethora of sophisticated sampling algorithms have been developed over the years, e.g. Metropolis-Hastings~\cite{metropolis1953equation} or Hamilton Monte Carlo~\cite{duane1987hybrid}.
However, most of these advanced sampling techniques are not designed to handle constrained spaces and it is not straightforward to apply these techniques to constrained spaces ---exceptions exist~\cite{betancourt2011nested,afshar2016closed}. 
A different road is to specifically design MC algorithms for constrained spaces. One such algorithm is VolEsti, which is presented in~\cite{emiris2014efficient,cousins2016practical}\footnote{An efficient \cpp implementation is available at \url{https://github.com/GeomScale/volume_approximation}.}. VolEsti is a Markov Chain Monte Carlo algorithm, in the family of \textit{hit-and-run} samplers~\cite{simonovits2003compute}, which is able of sampling from convex regions in high-dimensional spaces. Unlike~\cite{betancourt2011nested,afshar2016closed}, VolEsti is additionally capable of estimating the volume of the convex region, from which samples are drawn, up to a user-defined threshold by recursively constructing co-centric balls of diminishing radii, a technique introduced in~\cite{dyer1991random}. The complexity of \volesti is stated in the following theorem (cf. \cite[Theorem 2.1]{kannan1997random}):
\begin{theo} Given an $N-$dimensional convex body $K$, the precision parameter $\epsilon$, and the upper bound on the probability of error $\eta$, there is a (randomized) algorithm that returns a real number $\xi$ such that
\begin{talign}\label{eqn:error_volesti}
  (1-\epsilon)\xi < \wvol(K) < (1+\epsilon)\xi
\end{talign}
with probability at least $1-\eta$. The algorithm takes
\begin{talign}
  \mathcal{O}\Big( \frac{N^5}{\epsilon^2} (\ln^3{\frac{1}{\epsilon}})  (\ln{\frac{1}{\eta}})  (\ln^5{N}) \Big) = \mathcal{O}^*(N^5)
\end{talign}
oracle calls.

\end{theo}

\section{Monte Carlo Anti-Differentiation}

\subsection{Monte Carlo WMI}
The first step towards Monte Carlo anti-differentiation is to formally write down an MC approximation of the volume computation in Equation~\ref{eqn:volume_defin}, i.e. an MC estimate of the definite integral. Such an approximation was first given for WMI in~\cite{zuidbergdosmartires2018exactcompilation}, where the authors assumed that the weight function $\wweight$ is a probability density function defined over $\mathbb{R}^N$ and proposed an algorithm that samples values directly from the probability density. However, while sampling, constraints were not taken into account, which boils down to performing rejection sampling. 

 We give now the expression for the MC estimate of the weighted model integral when samples are drawn uniformly from points within convex polytopes. This might look like a step into the wrong direction at first. However, formulating the MC approximation of the weighted model integral in terms of uniform samples allows us to deploy the hit-and-run sampler of \volesti (cf. Section~\ref{sec:volesti}) to perform MC integration in high-dimensional spaces, thereby avoiding prohibitively high sample rejection rates.

\begin{theo}\label{theo:wmi_mc} (MC approximation of WMI) Let $\wsupport$ be an SMT(\lra) theory, $\wweight$ a weight function over the continuous variables $\xvars$. Then the volume computation $\wvol(\support,\weight | \xvars, \bvars)$ can be approximated by the following Monte Carlo estimate:
\begin{align}
    \wvol(\support,\wweight | \xvars, \bvars) \approx \sum_{\bvars}
 \frac{\polyvol^{\support}_{\bvars}}{N_{\bvars}} \sum_{i=1}^{N_{\bvars}}\ive{\support^\bvars(\xvars_i)}  \wweight(\xvars_i),& & 1{\ll} N_{\bvars} \label{eqn:mc_estimate_vol}
\end{align}
$N_{\bvars}$ is the number of samples uniformly drawn from the convex polytope induced by the linear constraints in $\ive{\support(\xvars_i, \bvars)}$ and $\polyvol^{\support}_{\bvars}$ denotes its volume.\\
{\bf Proof.} The proof is trivial and follows directly from the definition of Monte Carlo integration (cf.~\cite{weinzierl2000introduction}).\qed 
\end{theo}

\begin{theo}\label{theo:wmi_mc_error} (Error on MC estimate) Let $I= \int \ive{\support^\bvars(\xvars_i)}\wweight(\xvars_i) d\xvars$ and $N_{\bvars}{\rightarrow} \infty$.  The MC estimate of $I$   lies with probability $\frac{1}{\sqrt{2\pi}} \int_{-\epsilon}^{\epsilon} exp(-\frac{t^2}{2})dt$ within
\begin{equation}\label{eqn:error_mc_integral}
\Bigg[I-\epsilon\frac{ \sigma(\ive{\support^\bvars(\xvars_i)}\wweight(\xvars_i))}{\sqrt{N^\bvars}},
I+\epsilon\frac{\sigma(\ive{\support^\bvars(\xvars_i)}\wweight(\xvars_i))}{\sqrt{N^\bvars}}\Bigg]
\end{equation}
where $\sigma$ denotes the standard deviation and $N^\bvars$ is the number of samples (cf. \cite{weinzierl2000introduction}).
\end{theo}

\subsection{An Example}
Before formally introducing Monte Carlo anti-differentiation we first discuss MCAD in an example.

\begin{figure}[t!]
    \begin{center}
    \begin{minipage}[t]{0.4\linewidth}
            \begin{center}
            \includegraphics[width=\linewidth]{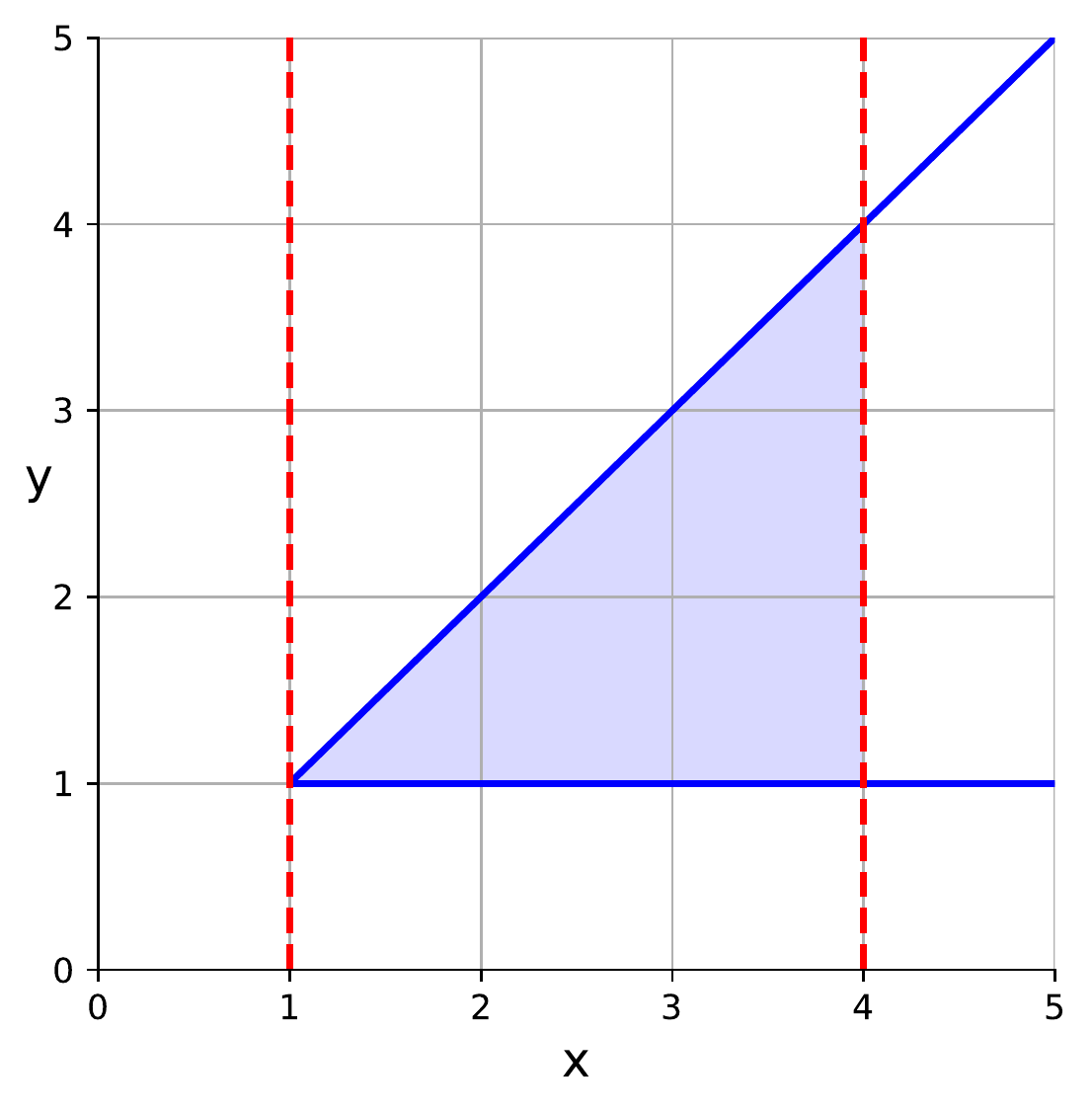}
            \end{center}
    \end{minipage}
    \begin{minipage}[t]{0.59\linewidth}
            \begin{center}
            \includegraphics[width=0.8\linewidth]{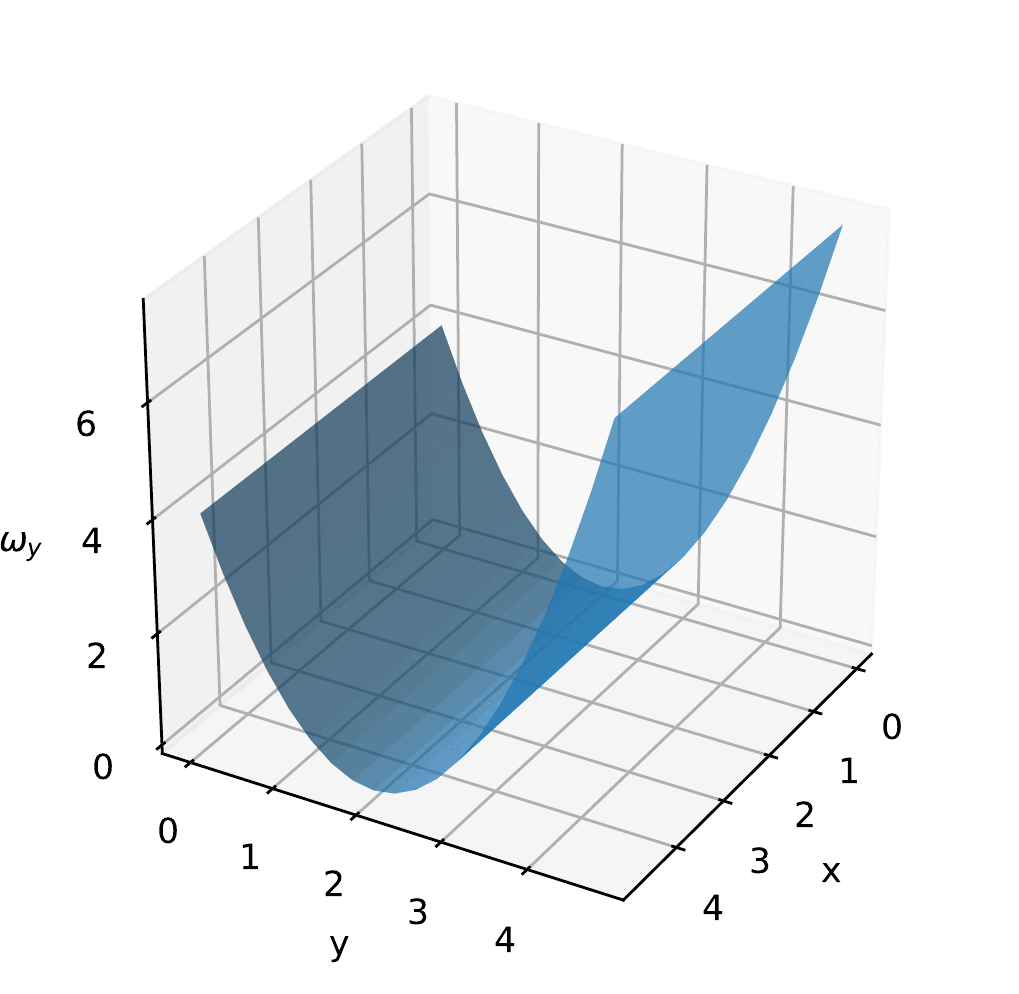}
            \end{center}
    \end{minipage}
    \caption{On the left is given a convex polytope (shaded in blue) constrained by the inequalities $(y{\geq} 1)$, $(x{\geq} y)$ (blue), and $(x{\geq}1)$, $(x{\leq} 4)$ (dotted red). On the right the weight function $\wweight(y)=(y-2)^2$ is shown.}%
    \label{figure:example_mcad}
    \end{center}
\end{figure}

\begin{figure}[t!]
    \begin{minipage}{.4\linewidth}
        \begin{center}
        \includegraphics[width=\textwidth]{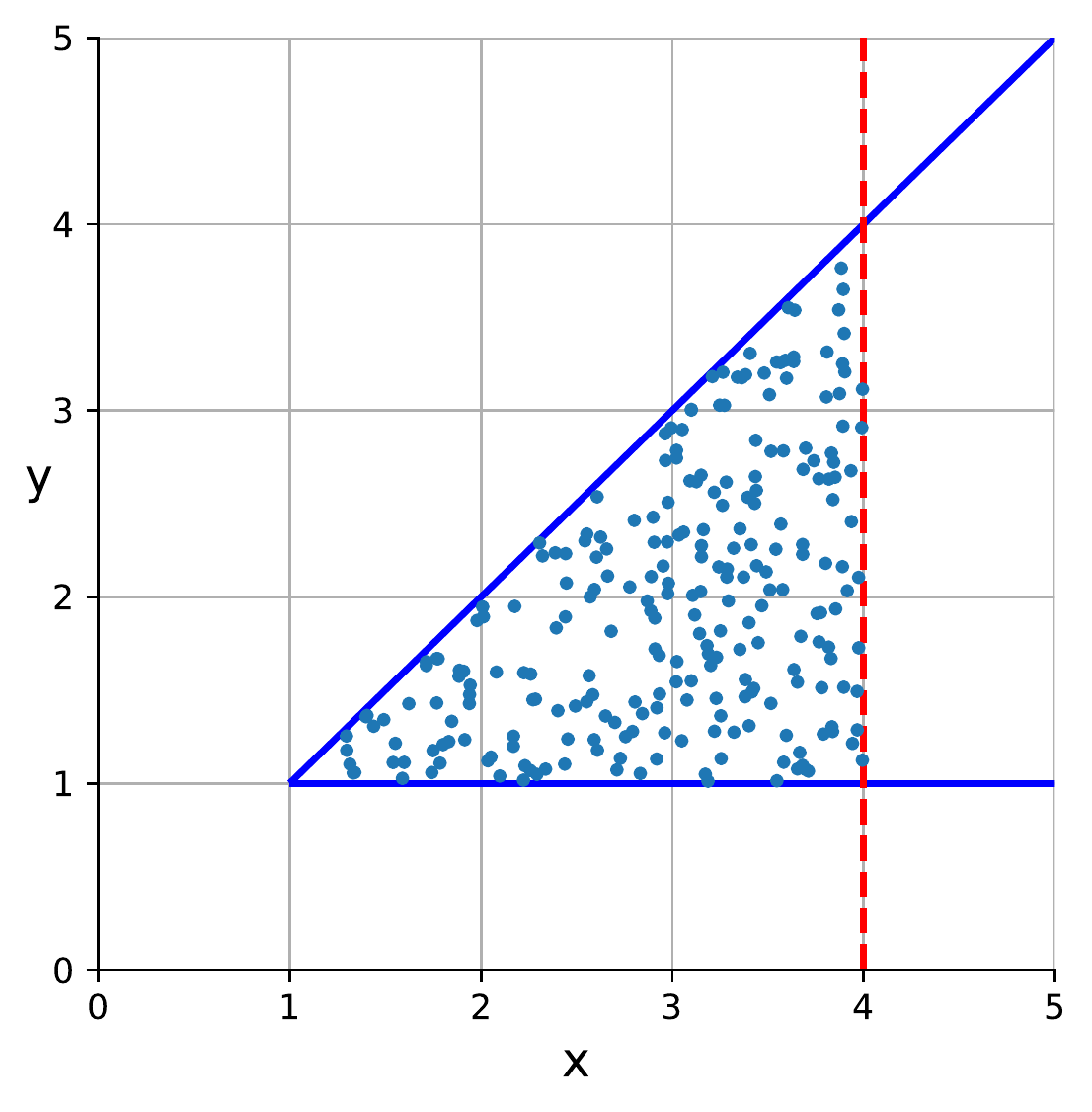}
        \end{center}
    \end{minipage}
    \begin{minipage}{.53\linewidth}
        \begin{center}
        \includegraphics[width=0.65\textwidth]{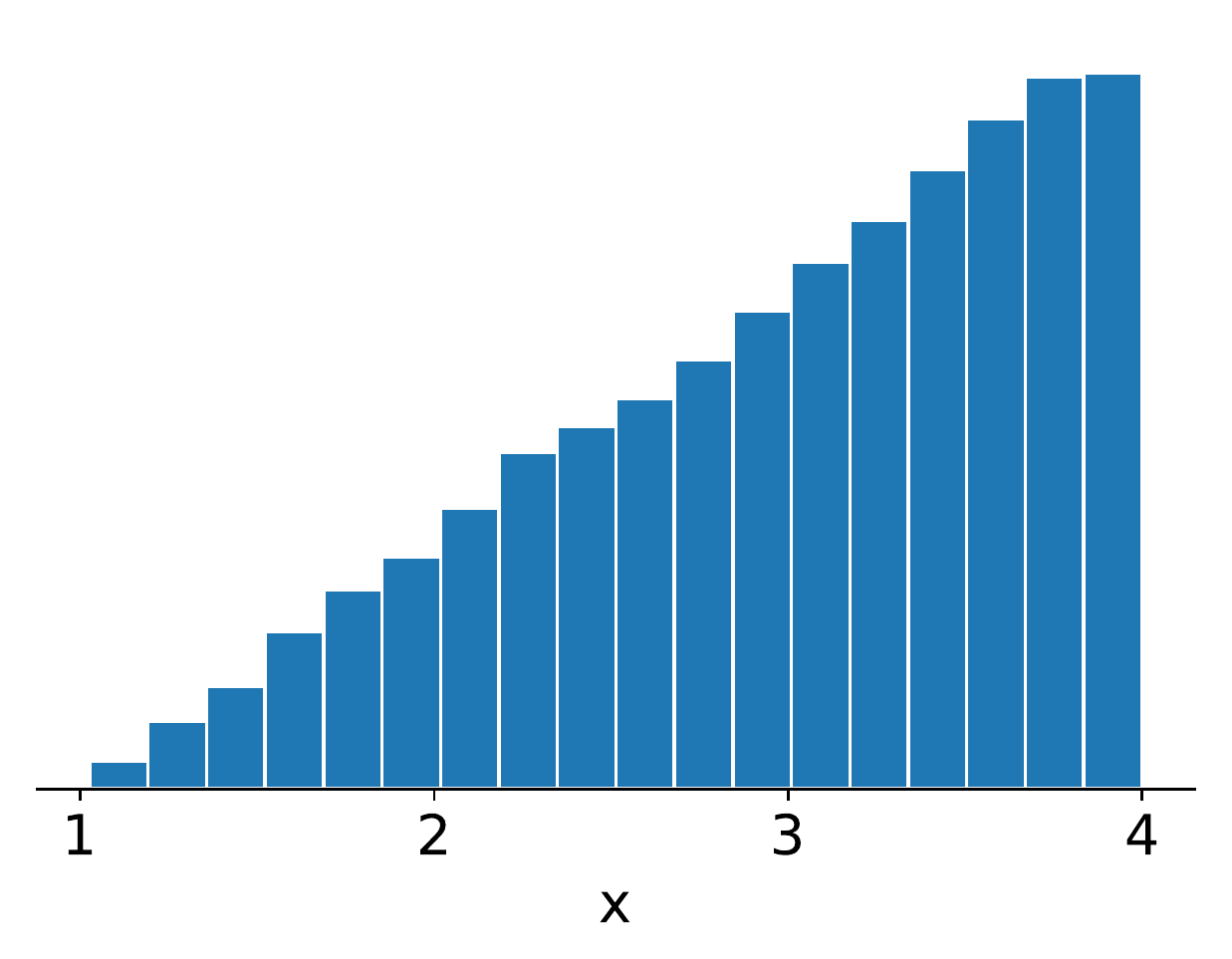}
        \includegraphics[width=0.65\textwidth]{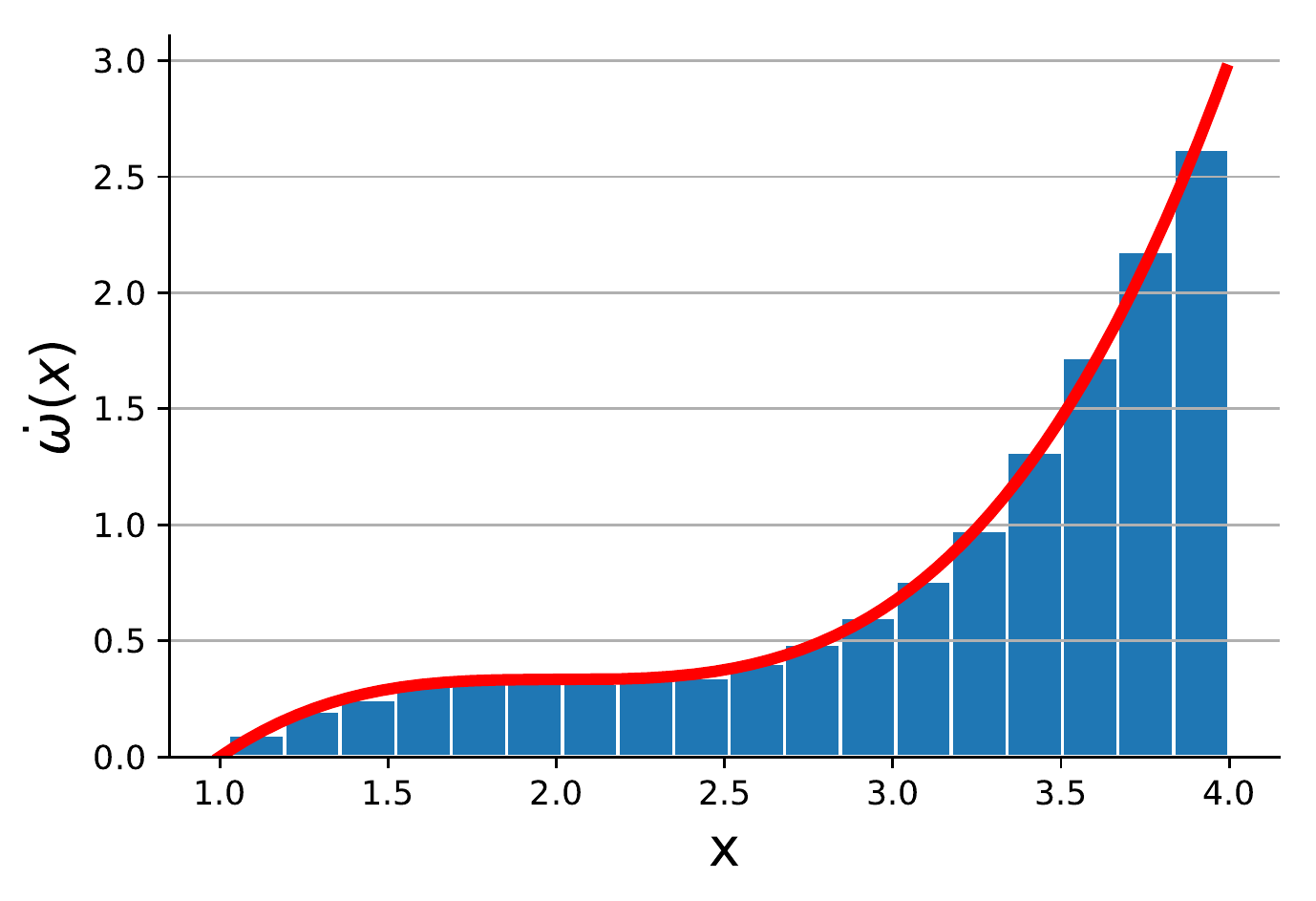}
        \end{center}
    \end{minipage}  
    \caption{The left plot shows samples drawn uniformly from within the convex polytope delimited by the constraintsy $y{\geq}1$, $x{\geq}y$ and  $x{\leq}4$. The histogram in the upper right shows the projection of the samples onto the $x$-axis and binned into $18$ bins of equal width. In the lower right is shown the piecewise constant function that approximates the anti-derivative of $y$ over the weight function $\wweight(y)$ constrained by the convex polytope. The anti-derivative is denoted by $\Psi(x)$. $\Psi(x)$ is obtained by weighting the samples with the weight function $\wweight(y)=(y-2)^2$ and estimating the density through a histogram. The plot does also show in red the (exact) symbolic integral: $\int_1^x (y-2)^2 dy = x^3-2x^2+4x-\nicefrac{7}{3}$.}
    \label{figure:example_samples2histogram}
\end{figure}

\begin{example}\label{example:mcad}
Consider the convex polytope given in Figure~\ref{figure:example_mcad} on the left and the weight function $\wweight(y)$ depicted in Figure~\ref{figure:example_mcad} on the right. We would like to compute the anti-derivative of the weight function with respect to $y$, which means that we need to integrate out $y$ taking into consideration the bounds imposed on $y$ ($y{\geq}1$ and $x{\geq}y$ (blue)). The problem is that theses bounds do not induce a convex polytope but  only a region in $\mathbb{R}^2$ and we are not able to deploy the sampling algorithm introduced in Section~\ref{sec:volesti}. To fix this, we take also into consideration the domain of the WMI problem on the $x$ variable --- $(x{\geq} 1)$ and $(x{\leq}4)$ (in dashed red). These four inequalities combined do now induce a convex polytope (shaded in blue).

With the convex poltyope at hand, we can now sample uniformly points from it. We then weight each sample with the weight produced by the weight function $\wweight(y)=(y-2)^2$. The Monte Carlo approximation of the anti-derivative is then obtained by estimating the density of the weighted samples. In Figure~\ref{figure:example_samples2histogram} we performed the density estimation using a histogram (bottom right).
\end{example}

Note that while we used a 1histogram, i.e. a piecewise constant function, to estimate the density for the anti-derivative in Figure~\ref{figure:example_samples2histogram}, this can be replaced with any density estimator, which we will discuss in the next subsection.

\subsection{MCAD}
We are now going to generalize Example~\ref{example:mcad} such that it fits the general setting of WMI over SMT(\lra) formulas.
\begin{theo}\label{theo:mcad}
Given is the $N$-dimensional convex poltyope $K(\xvars)$, which is defined by $M$ inequalities over the variables $\xvars$.
The Monte Carlo estimate $\tilde{\antid}$ of the anti-derivative $\antid$  of the weight function $\wweight(\xvars_I)$ with respect to the set of integration variables $\xvars_I{\subset} \xvars$, constrained by the convex polytope $K(\xvars)$ is given by:
\begin{align}
\hat{\Psi}=\ive{\land_{j}^{M^*} \support_j(\xvars_F, \xvars_C)} \xspace \Xi \Bigg( \Bigg\{ \xvars_C^i, \frac{\wvol({K^{CD}}) {\times}\wweight(\xvars_I^i)}{N} \Bigg\}_{i=1}^{N} \Bigg)
\end{align}
$\xvars_F {\subset} \xvars$, with $\xvars_F {\cap} \xvars_I {=} \emptyset$, are free variables that do not appear together with variables in $\xvars_I$ in any of the SMT(\lra) atoms. $\xvars_C{\subset} \xvars$, with $\xvars_F {\cap} \xvars_I {=} \emptyset$, are variables that are coupled to the variables in $\xvars_I$ by appearing together in at least one SMT(\lra) atom. $\Xi$ is a density estimator that estimates the density of the $N$ samples $\xvars_C^i$ weighted by $\wweight(\xvars_I^i)$, where the samples are drawn from the sub-polytope over the variables $\xvars_C{\cup}\xvars_I$ induced by the polytope $K(\xvars)$.
\\
{\bf Proof.} We start by writing the expression of the anti-derivate $\antid$ and expressing the polynomials $K(\xvars)$ as function of the $M$ SMT(\lra) atoms.
\begin{talign}
\antid 
&= \int K(\xvars)\wweight(\xvars_I)d\xvars_I\\
&= \int \ive{\land_{i=1}^{M} \support_i(\xvars)}\wweight(\xvars^*) d\xvars_I
\end{talign}
Now we separate the SMT(\lra) atoms by grouping together the $M^D$ atoms depending on $\xvars_I$ (indexed by $i$) and the $M^*$ atoms not depending on  $\xvars_I$ (index by $j$)
\begin{talign}
\antid 
&= \int \ive{\land_{j}^{M^*} \support_j(\xvars_F, \xvars_C)}
    \ive{\land_{i}^{M^D} \support_i(\xvars_C, \xvars_I)}
    \wweight(\xvars_I) d\xvars_I \\
&=\underbrace{\ive{\land_{j}^{M^*} \support_j(\xvars_F, \xvars_C)}}_{\eqqcolon \ive{\phi_{FC}}}
    \int \ive{\land_{i}^{M^D} \support_i(\xvars_C, \xvars_I)}
    \wweight(\xvars_I) d\xvars_I
\end{talign}
The Iverson brackets we would like to integrate over do not necessarily form a convex polytope, which would allow us to use the MCMC samples introduced in Section~\ref{sec:symbolic_integration}.  In order to obtain a convex polytope, we first explicitly write the lower and upper bounds on the variables in $\xvars_C$, $\support_L(\xvars_C)$ and $\support_U(\xvars_U)$ respectively,  and push them inside the integral.
\begin{talign}
\antid
=& \ive{\phi_{FC}}
    \ive{\support_{L}(\xvars_C)} \ive{\support_{U}(\xvars_C)}
    \int \ive{\land_{i}^{M^D} \support_i(\xvars_C, \xvars_I)}
    \wweight(\xvars_I) d\xvars_I\\
=& \ive{\phi_{FC}}
    \int
    \underbrace{
    \ive{\support_{L}(\xvars_C) }  \ive{\support_{U}(\xvars_C)}
    \ive{\land_{i}^{M^D} \support_i(\xvars_C, \xvars_I)}
    }_{\eqqcolon K^{CD}(\xvars_C,\xvars_I)}
    \wweight(\xvars_I) d\xvars_I \label{eqn:pushing_bounds_integral}
\end{talign}
Performing the integral results in a density dependent on the variables in $\xvars_C$:
\begin{talign}
\antid
&= \ive{\support_{FC}}
    \underbrace{
    \int K^{CD}(\xvars_C,\xvars_I)\wweight(\xvars_I)d\xvars_I
    }_{\eqqcolon \rho(\xvars_C)}
\end{talign}
We estimate the density by using $N$ samples drawn from $K^{CD}(\xvars_C, \xvars_D)$ which are weighted according to $\frac{\wvol({K^{CD}}) {\times}\wweight(\xvars_I)}{N}$:
\begin{align}
\hat{\rho}(\xvars_C)
&= \Xi \Bigg( \Bigg\{ \xvars_C^i, \frac{\wvol({K^{CD}}) {\times}\wweight(\xvars_I^i)}{N} \Bigg\}_{i=1}^{N} \Bigg)\label{eqn:density_esimtate_weighted_points}
\end{align}
where $\wvol(K^{CD})$ is the volume of the polytope $K^{CD}$ and  $\Xi$ a density estimator. The weights for the samples are obtained from the Monte Carlo estimate, cf. Equation~\ref{eqn:mc_estimate_vol}. $\xvars_C^i$ and $\xvars_I^i$ denote the $i$-th sample.
\qed
\end{theo}
Note that in the proof above we assumed that the upper and lower bounds on the variables in $\xvars_C$ were readily available. If this is not the case, they can be obtained by means of linear programming. Note also that adding the pushing the bounds on the variables $\xvars_C$ in Equation~\ref{eqn:pushing_bounds_integral} is reflected in Example~\ref{example:mcad} when adding the bounds on $x$ (in dashed red in Figure~\ref{figure:example_samples2histogram}) to the bounds on $y$. Furthermore, $\xvars_F{=}\emptyset$ in Example~\ref{example:mcad}. 

\subsection{Histograms as Density Estimator}
The perhaps simplest approach to estimate the density that produced a set of data points is by means of (multidimensional) histograms, i.e. by piecewise constant functions. In order to estimate the density in Equation~\ref{eqn:density_esimtate_weighted_points} we first need to obtain the domain on which we would like to estimate $\rho(\xvars_C)$. For each of the variables in $\xvars_C$ we take its lower and upper bound. The so obtained hyperrectangle $HR_\rho$ defines the bounds of the multivariate histogram. The hyperrectangle is then partitioned into a predefined number of bins $M_b$ of volume $v_b$ ($v_b$ depends on $M_b$ and the volume of $HR_\rho$). The estimate of the density for the $i$-th bin denoted by $b_i$ , with $1{\leq}i{\leq}M_b$ is given by:
\begin{align}
\hat{\rho}_H(\xvars_C \in b_i)
= \frac{\wvol(K^{CD})}{v_b N} \sum_j^{N} \bigg( \mathds{1}_{\xvars_C^j\in b_i} \times \wweight(\xvars_I^j) \bigg)
&& \forall i: 1{\leq}i{\leq}M_b
\end{align}
where $N$ is the number of samples which we use to estimate the density.

Even though using histograms as density estimators is straight forward they are not well suited for estimating high dimensional data. If we want to preserve the bin resolution, the number of bins we need to partition our space into grows exponentially with the number variables present in the set $\xvars_C$. A possible solution would be to represent the estimate the density $\rho(\xvars_C)$ by a lower dimensional representation instead of a piecewise constant function encoded through exponentially many bins, such as kernel density methods~\cite{rosenblatt1956remarks,parzen1962estimation}, density estimation trees~\cite{ram2011density}, or hybrid sum-product-networks~\cite{molina2018mixed}.

\paragraph{Error Bounds} We are now interested in the error that originates from approximating a function by a piecewise constant function. Therefore, we assume that the true density $\rho(\xvars_C)$ is the (midpoint) interpolant of the histogram $\hat{\rho}_H$ (we are not interested in the error introduced by the Monte Carlo approximation of $\rho$, this error is described in Theorem~\ref{theo:wmi_mc_error}). We first state a proposition, which we need to carry out the proof for the bounds on the error.

\begin{proposition} Let $U\subset \mathbb{R}^n$ be open, $f:U\rightarrow \mathbb{R}$ be differentiable, and the segment $[\bvec{a},\bvec{b}]$ joining $\bvec{a}$ to $\bvec{b}$ be contained in U. Then there exists $\bvec{c} \in [\bvec{a},\bvec{b}]$ such that
\begin{align}
    | f(\bvec{b})-f(\bvec{a}) |
    &\leq  \bigg( \sup_{\bvec{c}\in [\bvec{a},\bvec{b}]} \big| \nabla f(\bvec{c}) \big| \bigg) \cdot
    \vert \bvec{b}-\bvec{a}\vert
\end{align}
$|\cdot|$ denotes the $L^1$ norm.
\\{\bf Proof.} Follows trivially from the \textit{mean value theorem for multivariate functions}, see for example~\cite[Theorem 1.9.1]{hubbard2015vector}.
\end{proposition}

\begin{proposition}\label{cor:bound_mean_val}
Let $b_i\subset \mathbb{R}^n$ be the set of points in the $i$-th bin of the histogram $\hat{\rho}_H$, $f:b_i \rightarrow \mathbb{R}$ be a polynomial function, $d_b$ be the $L^1$ distance of the midpoint to one of the vertices of a bin. The error $\epsilon$ on the approximation of $f$ by $\hat{\rho}_H$ is bounded by:
\begin{align}
    \epsilon \leq d_b\max_{\xvars \in b_i} |\nabla f(\xvars)|
\end{align}
{\bf Proof.} The error on the histogram with regards to to the actual value of the polynomial $f$ for a given point $\xvars^* \in b_i$ is:
\begin{align}
    \epsilon(\xvars^*) &= |f(\xvars)- \hat{\rho}_H(\xvars) | 
\end{align}
We notice now that for every $\xvars\in b_i$ $\hat{\rho}_H(\xvars)=f(\xvars_{mp})$, where $\xvars_{mp}$ is the midpoint of the $i$-th bin. This gives us:
\begin{align}
    \epsilon(\xvars^*) &= |f(\xvars^*)- f(\xvars_{mp}) | &&\\
    \epsilon(\xvars^*) &\leq \bigg( \sup_{\xvars \in [\xvars^*,\xvars_{mp}]} \big| \nabla f(\xvars) \big| \bigg) \cdot
    \vert \xvars^*-\xvars_{mp}\vert &\text{using Proposition~\ref{cor:bound_mean_val}}&
\end{align}
The upper bounds for the error on the bin is the maximum value of $\epsilon(\xvars^*)$, i.e. $\epsilon=\max_{\xvars^*} \epsilon(\xvars^*)$. We then obtain:
\begin{align}
    \epsilon &\leq \max_{x^*} \Bigg[ \max_{\xvars \in [\xvars^*,\xvars_{mp}]}\bigg( \big| \nabla f(\xvars) \big| \bigg) \cdot
    \vert \xvars^*-\xvars_{mp}\vert \Bigg]
\end{align}
Where we also used the fact that for polynomials we can replace the supremum with a maximum.

Realizing that the maximal distance between the midpoint and any other point is the distance between the midpoint $\xvars_{mp}$ and a corner point $\xvars_{cp}$ of the hypercube finishes the proof:
\begin{align}
    \epsilon
    &\leq \max_{x^*} \Bigg[ \max_{\xvars \in [\xvars^*,\xvars_{mp}]}\bigg( \big| \nabla f(\xvars) \big| \bigg) \Bigg] \cdot
    \underbrace{\vert \xvars_{cp}-\xvars_{mp}\vert}_{=d_b}\\
    &\leq d_b \max_{\xvars \in b_i} \bigg( \big| \nabla f(\xvars) \big| \bigg)
\end{align}
\end{proposition}

To conclude this subsection, the density estimation introduces an additional error on the Monte Carlo anti-derivative (besides the errors on the Monte Carlo approximation itself (Equation~\ref{eqn:error_mc_integral}) and the error on the volume estimation (Equation~\ref{eqn:error_volesti}).

\section{\fmcad}\label{sec:mcad_algorithm}
Replacing the exact symbolic integration back-end with a Monte Carlo estimate gives the \fmcad algorithm. Structurally, the \fmcad follows the \fxsdd algorithm proposed in~\cite{kolb2019structure}, and also shown in Algorithm~\ref{alg:fac}. The sole difference to the original \fxsdd algorithm is that in the \fmcad algorithm the exact symbolic anti-differentiation is substituted by a Monte Carlo approximation, whose computation is given in Equation~\ref{theo:mcad}.

\begin{algorithm}[tb]
    \caption{Factorized Integration}\label{alg:fac}
    \begin{algorithmic}[1]
        \State world-weight $\wweight$
        \Procedure{$\fso$}{XSDD $D$, vars $\xvars$}
            \If{$\xvars = \emptyset$}
                \State \Return $\ive{D}$
            \ElsIf{$D$ is terminal}
            	\State \Return $\int \ive{D} \prod_{x \in \xvars} \wweight_x(x) d\xvars$
            \ElsIf{$D = \bigvee_{c} D_c$}
                \State \Return $\sum_{c} \fso(D_c, \mathbf{x}) \label{line:or}$
            \ElsIf{$D = D_1 \land D_2$}
                \State $\xvars_s = \xvars \cap \mathrm{vars}(D_1) \cap \mathrm{vars}(D_2)$ \label{alg:fac:vars_inter}
                \State $\xvars_1^*, \xvars_2^* = \mathrm{vars}(D_1) \setminus \xvars_s, \mathrm{vars}(D_2) \setminus \xvars_s$ \label{alg:fac:vars_diff}
                \State $r_1 = \fso(D_1, \xvars_1^* \cap \xvars)$ 
                \State $r_2 = \fso(D_2, \xvars_2^* \cap \xvars)$ 
                \State \Return $\int r_1 \cdot r_2 \cdot \prod_{x \in \xvars_s} \wweight_x(x) d\xvars_s \label{line:and}$
            \EndIf
        \EndProcedure
    \end{algorithmic}
\end{algorithm}

Crucial hyperparameters of \fmcad are the bin resolution of the histograms, which we use for the density estimation, and the number of samples used for the Monte Carlo integration. A further choice to be made is the method to be used to perform the  volume computation of convex polytopes, i.e. whether to perform this volume computation exactly or approximately.

In future work we would also like to perform a theoretical analysis of the \fmcad algorithm where we analyze the error propagation when evaluating a underlying arithmetic circuit. Such an analysis would follow ideas presented in~\cite{shah2019problp}.

\section{Experimental Evaluation}\label{sec:experiments}
The experimental study of \fmcad presented in this section answers the following questions. {\bf Q1} Can \fmcad exploit structure in highly-structured WMI problems and does \fmcad produce reliable MC approximations? {\bf Q2} How does \fmcad compare to naive rejection sampling and XSDD(Sampling)? {\bf Q3} How does \fmcad handle integrations in higher dimensional spaces?

\begin{figure}[ht!]
    \begin{center}
    \begin{minipage}[t]{0.45\linewidth}
            \begin{center}
            \includegraphics[width=\linewidth]{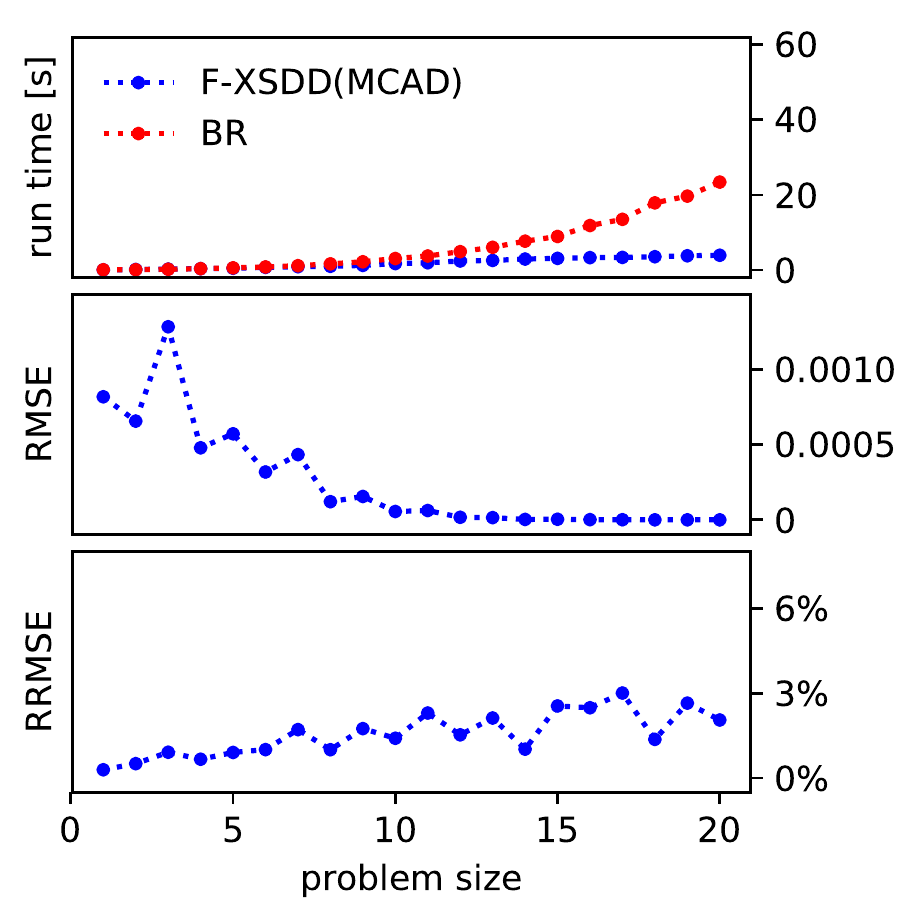}
            \subcaption{XOR($N$)}
            \label{figure:xor_xadd}
            \end{center}
    \end{minipage}
    \begin{minipage}[t]{0.45\linewidth}
            \begin{center}
            \includegraphics[width=\linewidth]{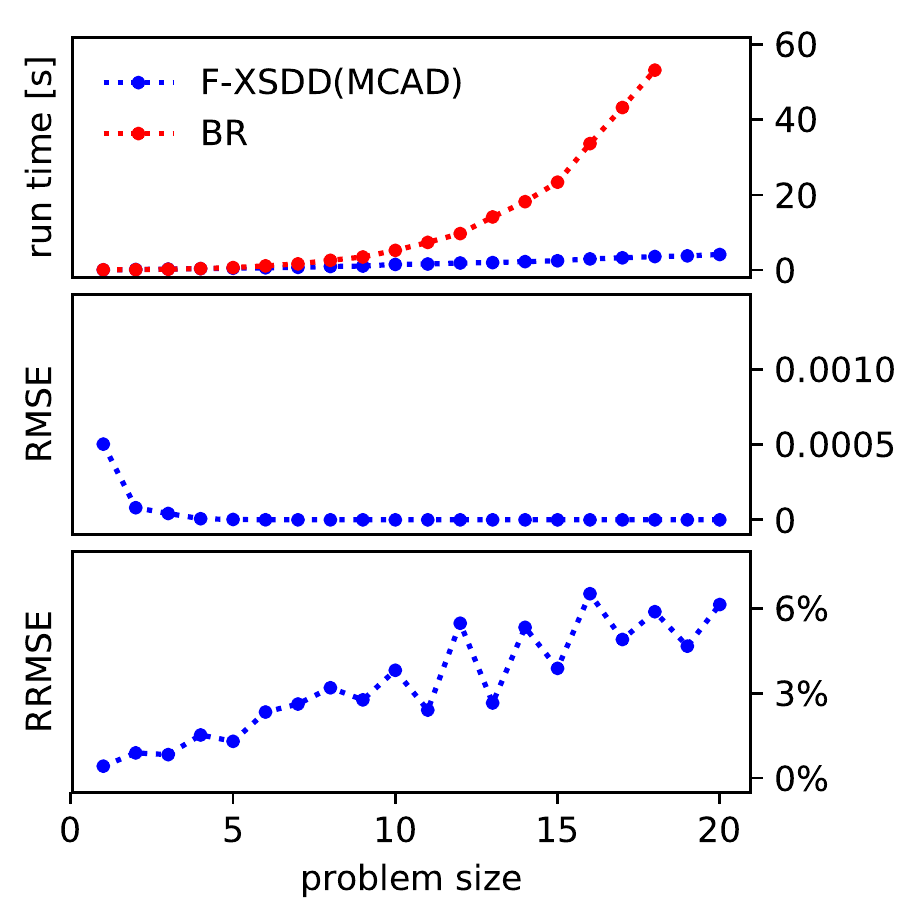}
            \subcaption{XOR($x_i^2$,$N$)}
            \label{figure:xor_x2i}
            \end{center}
    \end{minipage}
    \caption{Plotted are the run time, the root mean squared error (RMSE), and the relative RMSE (RRMSE) against the problem size $N$. \fxsdd used $50000$ samples per integration and used for the volume computation of convex polytopes the symbolic inference engine of PSI~\cite{gehr2016psi}. Run times, RMSE, and RRMSE were obtained over 10 runs.}%
    \label{figure:xor_xor_x2i_xadd}
    \end{center}
\end{figure}

In order to answer {\bf Q1} we compare \fxsdd to the BR algorithm of~\cite{kolb2018efficient} on the XOR($N$) benchmark (Figure~\ref{figure:xor_xadd}) ---  $N$ denotes the variable problem size. The XOR(N) benchmark is a highly structured synthetic problem, which most state-of-the-art solvers struggle with~\cite{kolb2019structure} and unless the BR algorithm is deployed, they exhibit an exponential dependency of the run time on the problem size. Besides the comparison on the XOR(N) benchmark we compare also the \fxsdd and BR algorithms on a variation of the XOR($N$) benchmark. Instead of having a constant weight function of one in the XOR($N$) benchmark, we use now a weight function $\weight=\prod_i x_i^2$ where the $x_i$ are the continuous variables present in the benchmark. We refer to this variation as XOR($x_i^2$,$N$)  

In Figure~\ref{figure:xor_xor_x2i_xadd} we see that the run time of \fxsdd does not exhibit the exponential growth of the run time with increasing problem size. This shows that \fxsdd is, just as BR, capable of exploiting the structure present by performing MC approximations of indefinite integrals. Furthermore, we observe that the run time increases more rapidly for the BR algorithm than for \fxsdd in both plots (being more prominent in Figure~\ref{figure:xor_x2i}). This is because symbolic integration of the symbolic inference engine in the BR algorithm starts slowing down the run time for larger problem sizes. Analyzing the the root mean squared error (RMSE) and the relative RMSE (RRMSE) we see that the solutions produced by \fxsdd are meaningful approximations. 

To answer {\bf Q2} we compare, on the one hand, \fxsdd to naive rejection sampling, which means that we sample uniformly from the problem domain (which is a hyperrectangle) and reject the samples that do not satisfy the SMT(\lra) constraints of the problem. On the other hand, we compare to XSDD(Sampling), an XSDD based algorithm, which collects all convex polytopes separately, finds bounding hypercubes for these and performs rejection sampling in these hypercubes. The comparisons are made for the XOR($N$) benchmark and the Mutex($N$) benchmark, cf.~\cite{kolb2018efficient}.

\begin{figure}[t!]
    \begin{center}
    \begin{minipage}[t]{0.45\linewidth}
            \begin{center}
            \includegraphics[width=\linewidth]{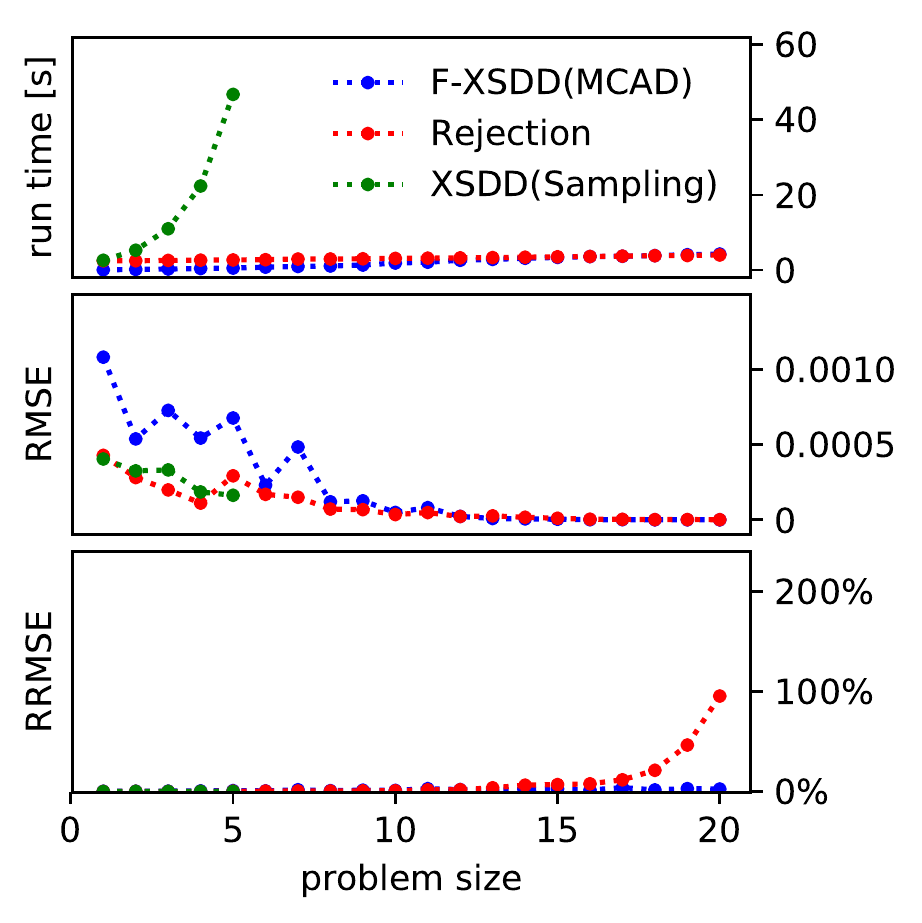}
            \subcaption{XOR(N)}
            \label{figure:xor_rejection}
            \end{center}
    \end{minipage}
    \begin{minipage}[t]{0.45\linewidth}
            \begin{center}
            \includegraphics[width=\linewidth]{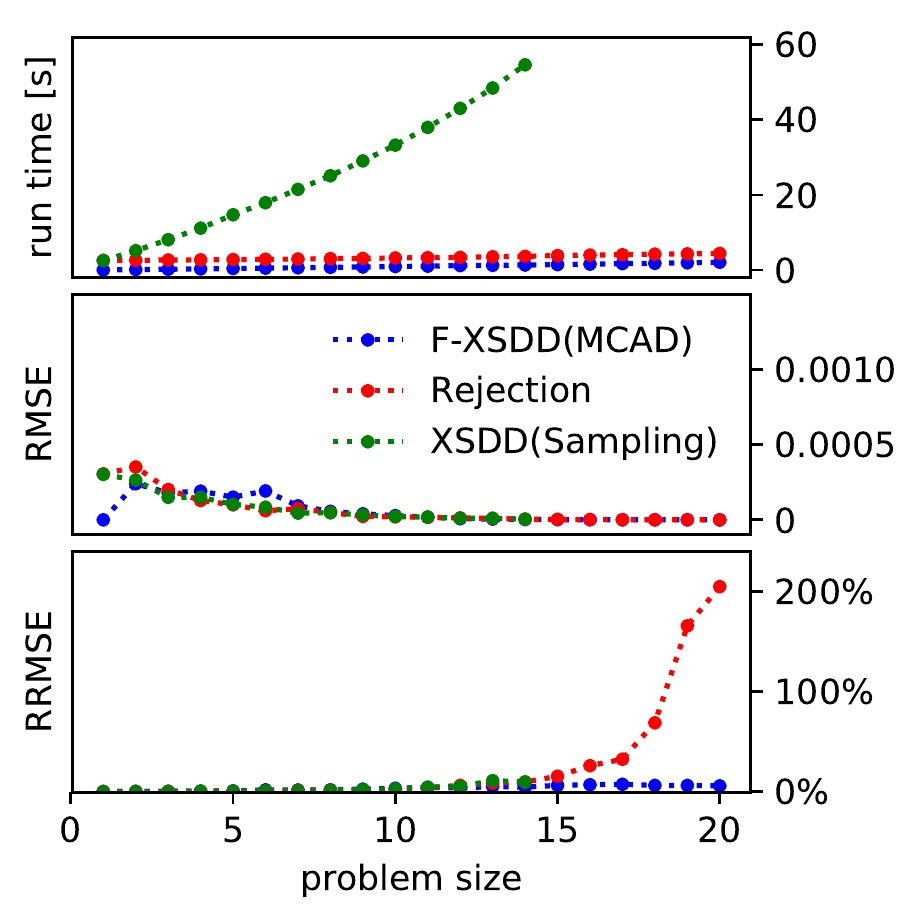}
            \subcaption{Mutex(N)}
            \label{figure:mutex}
            \end{center}
    \end{minipage}
    \caption{\fxsdd used $50000$ samples per integration and used for the volume computation of convex polytopes the symbolic inference engine of PSI. XSDD(Sampling) and naive rejetion sampling use $1.5 \times 10^6$ samples per integration. Run times, RMSE, and RRMSE were obtained over 10 runs.}%
    \label{figure:xor_mutex_rejection}
    \end{center}
\end{figure}

In the plots (Figure~\ref{figure:xor_mutex_rejection}), we observe the drawback of collecting convex polytopes: the number of integrations becomes prohibitively high and hurts the run time of the XSDD(Sampling) algorithm. For the XOR($N$) benchmark, an additional factor plays into the run time: the number of convex polytopes to be collected grows exponential with the problem size. We also see that for the naive rejection sampling approach the relative RMSE grows drastically for large problem sizes, which means that too many samples are being rejected. \fxsdd does not suffer from the drawbacks in run time of the XSDD(Sampling) algorithm nor from the sharp drop in accuracy of the naive rejection sampler.

We tackle {\bf Q3} by investigating a variation of the Dual benchmark~\cite{kolb2019structure}, which aims to show the benefit of factorized solving. We dub this variation M-ual~(M-ual($x^2_i,N,M$)), which is specified as follows:
\begin{talign}
\support&= \bigvee_{i=1}^N \big[ \sum_{k=1}^M (x_{ik}{\leq}0)\big]\nonumber \\
&\land \bigwedge_{i=1}^N\big[(\sum_{k=1, k\neq i}^M x_{ik}{\leq}0) \lor  {\lnot}(\sum_{k=1, k\neq i}^M x_{ik}{\leq}0)\big]\nonumber
\end{talign}
with domain $\bigwedge_{k=1}^M\bigwedge_{i=1}^N(-1{\leq} x_{ik}{\leq} 1)$ and $\weight {=} \prod_{i=1}^N \prod_{k=1, k\neq i}^M x_{ik}^2$. It is easy to see that by increasing $N$ and $M$ simultaneously a high-dimensional problem is created. For instance the problem M-ual($x_i^2, 10,10$) is 100-dimensional.

As for the experiment itself we compared variation of the \fmcad algorithm where we used different algorithms to compute the volume of the polytopes over which integrations are performed (cf. Equation. For completeness we also included the naive rejection sampling algorithm in the comparison.

\begin{figure}[t!]
    \begin{center}
    \begin{minipage}[t]{0.45\linewidth}
            \begin{center}
            \includegraphics[width=\linewidth]{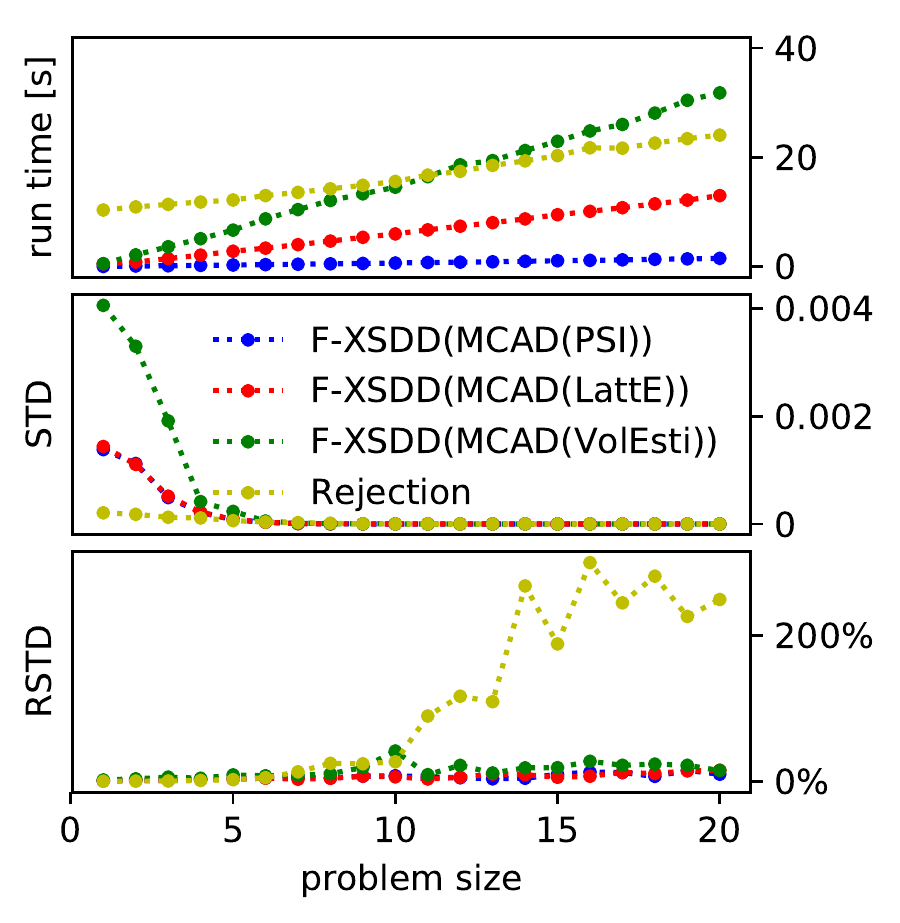}
            \subcaption{M-ual($x_i^2$,N,2)}
            \label{figure:dual_2xi}
            \end{center}
    \end{minipage}
    \begin{minipage}[t]{0.48\linewidth}
            \begin{center}
            \includegraphics[width=\linewidth]{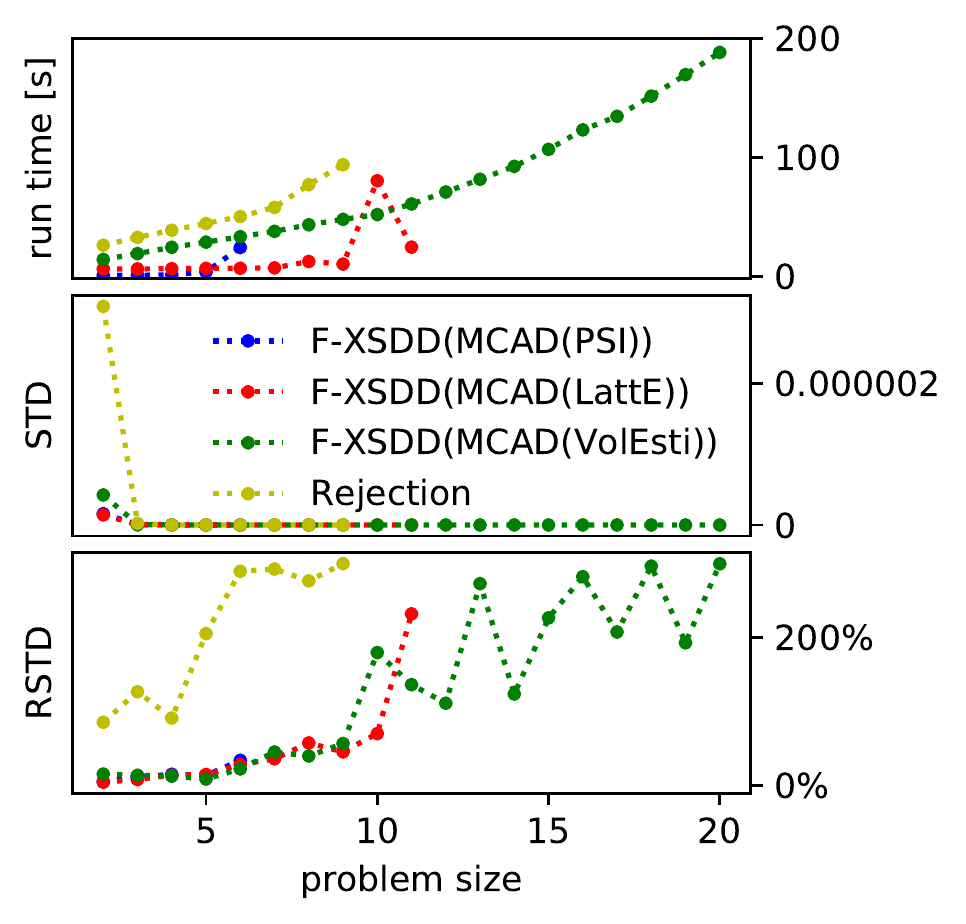}
            \subcaption{M-ual($x_i^2,10,M$)}
            \label{figure:M_uple_10}
            \end{center}
    \end{minipage}
    \caption{\fxsdd used $50000$ samples per integration and used for the volume computation of convex polytopes PSI, LattE or the Volesti (indicated in the legend). The naive rejection sampling algorithm used $6\times 10^6$ samples per integration for M-ual($x_i^2$,N,2) and $10^7$ samples for M-ual($x_i^2$,10,M). Note that contrary to the previous plots, we give now the standard deviation (STD) and the relative STD (RSTD) instead of the RMSE and RRMSR. Run times, STD, and RSTD were obtained over 10 runs. Missing data points for all but F-XSDD(MCAD(VolEsti)) algorithm are due to time outs.}%
    \label{figure:M_uple}
    \end{center}
\end{figure}

In Figure~\ref{figure:dual_2xi} we see that for comparable run times the rejection sampling algorithm's accuracy breaks down for higher dimensional spaces --- the space is $\sim$30-40 dimensional but the \fxsdd algorithms perform integrations merely in 2-dimensions due to the factorizability of the problem. In the run times, we see that for such low dimensional spaces of integration the exact methods (FXSDD(MCAD(PSI)) and FXSDD(MCAD(LattE)) outperform the approximate method (FXSDD(MCAD(VolEsti)). (Here we refer to to the computation of the volume of the polytopes as exact and approximate, and not to the integration itself.) We observe the opposite effect when tackling higher dimensional problems~(Figure~\ref{figure:M_uple_10}, where the exact algorithms time out for $M=6$ (FXSDD(MCAD(PSI)) and $M=11$ FXSDD(MCAD(LattE)). The naive rejection sampling algorithm timed out at $M=9$ and exhibited a considerably higher relative standard deviation, when compared to the \fxsdd algorithms.

\section{Conclusions and Future Work}
We developed the concept of Monte Carlo anti-differentiation, and proposed a method for performing an MC approximation of an indefinite integral. Based on Monte Carlo anti-differentiation we enriched the \fxsdd family of algorithms with \fmcad. \fmcad is based on a hit-and-run sampler, which is used to approximate the anti-derivative. As such, \fmcad is the first inference algorithm for WMI that performs Monte Carlo integration while exploiting structure present in WMI problems, and avoiding prohibitively high sample rejection rates. 
Even though the sample rejection rate is zero, the integration algorithm could still be improved by, instead of drawing samples uniformly, drawing them from the integrand directly. Deploying more sophisticated MC algorithms might lead to lower variance in the approximation of the integrals. Similarly, it would also be interesting to investigate approximate integration methods for WMI other than MC integration that circumvent the curse of dimensionality, such as sparse grids~\cite{bungartz2004sparse} and Bayesian quadrature~\cite{briol2015frank}.

\section*{Acknowledgements}
This work has received funding from ERC AdG SYNTH(694980). Samuel Kolb is supported by the Research Foundation-Flanders (FWO). Pedro Zuidberg Dos Martires is supported by Research Foundation-Flanders (FWO) and Special Research Fund of the KU Leuven (BOF). The authors would like to thank Vissarion Fisikopoulos for his help with the \volesti library and giving valuable feedback on the paper draft, and Luc De Raedt for commenting on early iterations of the paper.

\bibliography{references}
\bibliographystyle{aaai}

\end{document}